\documentclass[fleqn,11pt]{llncs}

\usepackage{url}
\usepackage{amsmath}
\usepackage{amssymb}
\usepackage{numprint}
\usepackage{todonotes}
\usepackage[utf8]{inputenc} 

\title{Feature extraction with Spectral Clustering for Gene Function
	Prediction using Hierarchical Multi-label Classification}

\author{
	Miguel Romero \and
	Oscar Ram\'irez \and
	Jorge Finke \and
	Camilo Rocha
}

\institute{
	Department of Electronics and Computer Science \\
	Pontificia Universidad Javeriana, Cali, Colombia
	\email{\{miguelangel.romero,jfinke,camilo.rocha\}@javerianacali.edu.co}
}

\begin{document}

\maketitle
	
\begin{abstract}
	Gene annotation addresses the problem of predicting unknown
	associations between gene and functions (e.g., biological processes)
	of a specific organism. Despite recent advances, the cost and time
	demanded by annotation procedures that rely largely on in vivo
	biological experiments remain prohibitively high. 
	This paper presents a novel in silico approach for to the annotation
	problem that combines cluster analysis and hierarchical multi-label
	classification (HMC). The approach uses spectral clustering to
	extract new features from the gene co-expression network (GCN) and
	enrich the prediction task. HMC is used to build multiple estimators
	that consider the hierarchical structure of gene functions. 
	The proposed approach is applied to a case study on \textit{Zea
		mays}, one of the most dominant and productive crops in the world. 
	The results illustrate how in silico approaches are key to reduce the
	time and costs of gene annotation. More specifically, they highlight
	the importance of: (i) building new features that represent the
	structure of gene relationships in GCNs to annotate genes; and (ii)
	taking into account the structure of biological processes to obtain
	consistent predictions.
\end{abstract}

\section{Introduction}
\label{sec:intro}

Identifying the association of genes to functions is key to gain
insight into how genomes serve as blueprints for life, e.g., to
develop treatments for specific conditions or enhance tolerance to
environmental
stresses~\cite{rust-annoteq-2002,vandepoele-arabidopsis-2009,yandell-annot-2012}. Numerous
studies have used co-expression data to predict specific biological
functions and
processes~\cite{oti-coexp-2008,romero-gene-2020,stuart-coexp-2003,vandam-disease-2017}. Intuitively,
genes are reported to co-express whenever they are simultaneously active,
which suggests that they are associated to one or more common
biological processes.

Under this hypothesis, characterizing gene interactions as a gene
co-expression network (GCN) may assist to identify unknown functional
annotations in a genome. Co-expression networks are generally
represented as undirected weighted graphs, where vertices denote genes
and weighted edges indicate the strength of the co-expression between
two genes. A detailed analysis of the structure and distribution of
gene relationships in GCNs provides additional clues that facilitate
the prediction of gene functions~\cite{valentini-tpr-2009}.

However, the cost and time requirements to annotate genes using in
vivo biological experimentation remains prohibitively
high~\cite{cho-diff-2015,zhou-insilicoannot-2005}. To overcome this
limitation, hybrid approaches that integrate existing knowledge of
gene-function associations and in silico methods have been
proposed~\cite{cho-functanalysis-2016,deng-prot-2003,luo-matrix-2007,romero-hmc-2022}. 
While they have shown great promise, given the extreme combinatorial
nature of the problem, annotating genes in an efficient manner remains
an open challenge.

Functional annotations are defined by the Gene Ontology (GO), which
contains three main types of annotations: biological processes,
molecular functions, and cellular component~\cite{go-go-2019}. These
annotations, commonly known as GO terms, are structured in a hierarchy
and defined as a directed acyclic graph (DAG). Gene annotation
approaches generally ignore the relationships among biological
processes, even though these relationships are key to improve the
accuracy and avoid inconsistency in predictions. 
A prediction is said to be inconsistent w.r.t. the GO hierarchy when a
gene is inferred to have a particular function $a$, but it is not
inferred to have all ancestor of $a$. In other words, an inconsistent
prediction states that the prediction does not satisfy the ancestral
relations between GO terms. Satisfying ancestral constraints is often
referred to as the \textit{true-path rule} in
GO~\cite{valentini-tpr-2009,ashburner-go-2000} and as the
\textit{hierarchical constraint} in HMC~\cite{vens-hmc-2008}.

This paper presents a feature extraction approach for in silico
annotation of genes. It follows a network-based approximation that
uses cluster analysis and hierarchical multi-label classification
(HMC) for building a predictor that assigns functions to genes
satisfying the true-path rule. Cluster analysis plays the role of
enriching the information available for predicting gene-function
associations by extracting new features that represent structural
properties of the GCN. That is, co-expression relations are used to
identify gene clusters that ultimately help in associating functions
to genes (i.e., guilt by association, see~\cite{petsko-gba-2009}). It
has been shown in~\cite{romero-clust-2022} that new features built
from the GCN and associations between genes and functions with the
spectral clustering algorithm are key to improve the prediction
performance in the gene annotation problem. The results
in~\cite{romero-clust-2022} show that using other features associated
to structural properties of the GCN and gene functional information
lead to lower performance.

Furthermore, the extracted features are filtered (using SHAP) based on
their impact in the prediction task and HMC is used to predict
gene-function associations that take into account the relations
between biological functions. The proposed approach illustrates how
the performance of gene annotation is improved by combining: (i) new
information extracted from the GCN; and (ii) classification methods
that consider the relation between gene functions.

This approach is applied to a case study on \textit{Zea mays}, one of
the most dominant and productive crops. \textit{Zea mays} serves a
variety of purposes, including animal feed and derivatives for human
consumption and ethanol~\cite{zhou-gcn-2020}.  The co-expression
information used in the study is imported from the ATTED-II
database~\cite{obayashi-atted2018-2018}. The resulting GCN, modeled as
a weighted graph, comprises \numprint{26131} vertices (i.e., genes)
and \numprint{44621533} edges. The functional information (i.e., known
gene-function associations) is taken from DAVID Bioinformatics
Resources~\cite{huang-david-2009}. It contains a total of
\numprint{255865} annotations of biological processes for maize, i.e.,
pathways to which a gene contributes. The results highlight the
importance of extracted features that represent structural properties
of the GCN and the hierarchical structure of biological processes with
HMC to improve prediction performance. Ultimately, the results provide
experimental (in silico) evidence that the proposed approach is a
viable and promising approximation to gene function prediction.

This paper is a significant extended version
of~\cite{romero-clust-2022} that:

\begin{itemize}
	\item Addresses the gene function prediction as a hierarchical
	multi-label classification problem by considering the structure of
	gene functions. That is ancestral relationships are represented as a
	DAG~\cite{go-go-2019}.
	
	\item Analyzes a larger functional database for the case study of
	maize. The number of genes associated to at least one function
	increased from \numprint{5361} to \numprint{10049}. The new dataset
	consists of \numprint{255865} associations between genes and
	functions, and \numprint{7021} relations between functions.
	
	\item Concludes that the ancestral relations between functions and
	the features extracted from the GCN improve the prediction
	performance in the gene function prediction task when addressed as a
	hierarchical multi-label classification problem.
\end{itemize}

The remainder of the paper is organized as follows.
Section~\ref{sec:prelim} reviews some preliminaries.
Section~\ref{sec:method-feat} introduces the approach to extract
features from the gene co-expression network using cluster analysis.
The proposed approach to predict gene functions, based on hierarchical
multi-label classification is presented in
Section~\ref{sec:method-pred}. Section~\ref{sec:case} presents the
case study for the \textit{Zea mays} species. Finally,
Section~\ref{sec:concl} draws some concluding remarks and future
research directions.

\section{Preliminaries}
\label{sec:prelim}

This section presents preliminaries on spectral clustering, gene
co-expression networks, gene function prediction, hierarchical
multi-label classification, and SHAP feature contribution.

\subsection{Spectral clustering}

The aim of applying cluster analysis on a network is to identify groups of
vertices sharing a (parametric) notion of
similarity~\cite{yu-spec-2003,rodriguez-clust-2019}. Usually, distance
or centrality metrics are used for clustering.
Spectral clustering is a clustering method with foundations in
algebraic graph theory~\cite{jia-spec-2014}. It has been shown that
spectral clustering has better overall performance across different
areas of applications~\cite{murugesan-clust-2021}. Given a
graph $G$, the spectral clustering decomposition of $G$ can be
represented by the equation $\mathbf{L} = \mathbf{D} - \mathbf{A}$,
where $\mathbf{L}$ is the Laplacian, $\mathbf{D}$ is the degree (i.e.,
a diagonal matrix with the number of edges incident to each node), and
$\mathbf{A}$ the adjacency matrices of $G$. Spectral clustering uses,
say, the $n$ eigenvectors associated to the $n$ smallest nonzero
eigenvalues of $\mathbf{L}$. In this way, each node of the graph gets
a coordinate in $\mathbb{R}^n$. The resulting collection of
eigenvectors serve as input to a clustering algorithm (e.g., k-means)
that groups the nodes in $n$ clusters.

\subsection{Gene Co-expression Network}

A gene co-expression network (GCN) is represented as an undirected
graph where each vertex represents a gene and each edge the level
of co-expression between two genes.

\begin{definition}\label{def.network.coexp}
	Let $V$ be a set of genes, $E$ a set of edges that connect pairs of
	genes, and $w:E \to \mathbb{R}_{\geq 0}$ a weight function. A
	\emph{(weighted) gene co-expression network} is a weighted graph $G
	= (V, E, w)$.
\end{definition}

The set of genes $V$ in a co-expression network is particular to the
genome under study. The correlation of expression profiles between
each pair of genes is measured, commonly, using the Pearson
correlation coefficient. Every pair of genes is assigned and ranked
according to a relationship measure, and a threshold is used as a
cut-off value to determine $E$. The weight function $w$ denotes the
strength of the co-expression between each pair of genes in $V$. For
example, in the ATTED-II database, the co-expression relation between
any pair of genes is measured as a $z$-score expressed as a function
of the co-expression index LS (Logit
Score)~\cite{obayashi-atted2018-2018,obayashi-coxpresdb-2011}.

\subsection{Gene Function Prediction}

In an annotated gene co-expression network, each gene is associated
with the collection of biological functions to which it is related
(e.g., through in vivo experiments).

\begin{definition}\label{def.network.ancoexp}
	Let $A$ be a set of biological functions. An \emph{annotated gene
		co-expression network} is a gene co-expression network $G = (V, E,
	w)$ complemented with an annotation function $\phi : V \to 2^{A}$.
\end{definition}

The problem of predicting gene functions can be explained as
follows. Given an annotated co-expression network $G = (V, E, w)$
with annotation function $\phi$, the goal is to use the information
represented by $\phi$, together with additional information (e.g.,
features of $G$), to obtain a function $\psi: V \to 2^A$ that
extends $\phi$. Associations between genes and functions not present
in $\phi$ have either not been found through in vivo experiments, or
do not exist in a biological sense. The new associations identified
by $\psi$ are a suggestion of functions that need to be verified
through in vivo experiments. The function $\psi$ can be built from a
predictor of gene functions, e.g., based on a supervised machine
learning model. 

\subsection{Hierarchical Multi-label Classification}

\textit{Node classification} refers to the task of predicting a node
class for an input data based on the information of other nodes in the
network~\cite{bhagat-nodecl-2011}. In general, node classification
problems can be categorize into three different types: \textit{binary
	classification} refers to predict one attribute (target) with two
classes (for example, positive and negative)~\cite{khan-binary-2010};
\textit{multi-class classification} refers to the case where the
attribute to be predicted has more than two classes and are mutually
exclusive (for example, the brand of a
car)~\cite{mills-multiclass-2021}; and \textit{multi-label
	classification} refers to predicting an attribute with at least two
classes, but where an instance could be associated to more than one
class (for example, the gene function prediction
problem)~\cite{xu-multi-2020}.

Although the aforementioned prediction methods are frequently used,
they do not consider hierarchical relations between classes. For such
scenarios, hierarchical multi-label classification (HMC) addresses the
task of structured output prediction where the classes are organized
into a hierarchy and an instance may belong to multiple classes. In
many problems, such as gene function prediction, classes inherently
satisfy these conditions~\cite{levatic-importance-2015}. Authors
in~\cite{silla-hierarchy-2011} expose that there are two types of
methods to explore the hierarchical structure. First, \textit{top-down
	or local classifiers} refer to partially predict the classes in the
hierarchy from the top to the bottom. Second, \textit{big-bang or
	global classifiers} refer to use a single classifier that considers
the entire hierarchy at once.

Classifiers that ignore the class relationships, by predicting only
the leaf classes in the hierarchy or predicting each class
independently, often lead to \emph{inconsistent predictions}. This
refers to the fact that a node is inferred to have a particular class
$a$, but the outcome of the classifier fails to infer the node's
association to all ancestor classes of $a$ in the hierarchy. In other
words, an inconsistent prediction states that the prediction does not
satisfy the hierarchy for some class $a$. Satisfying ancestral
constraints is often referred to as the \textit{true-path rule} in
GO~\cite{valentini-tpr-2009,ashburner-go-2000} and as the
\textit{hierarchical constraint} in HMC~\cite{vens-hmc-2008}.

\begin{figure}
	\centering
	\includegraphics[width=0.95\linewidth]{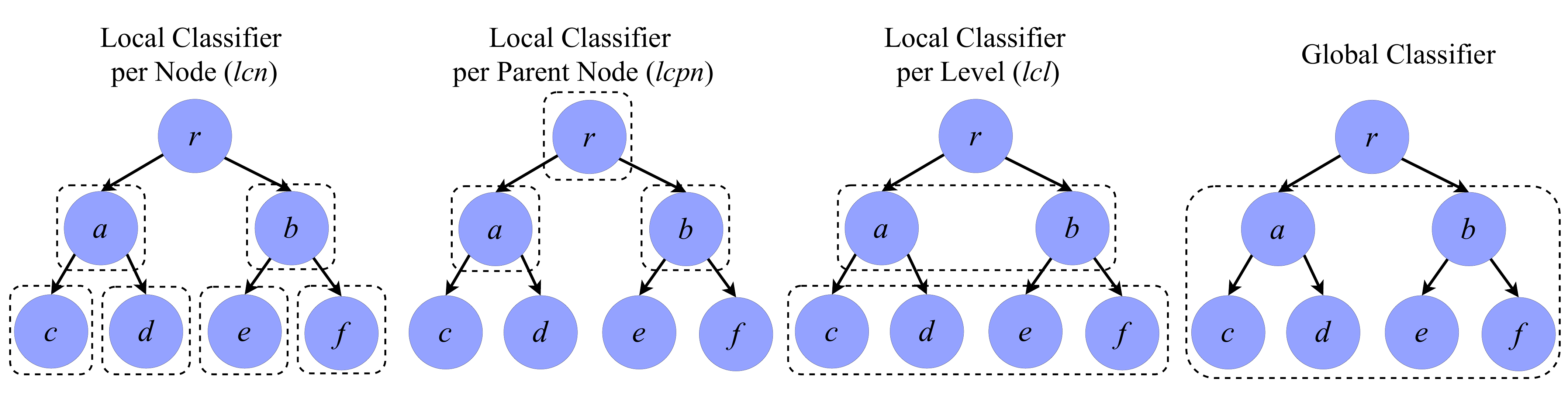}
	\caption{Example of global and local methods for hierarchical
		multi-label classification. Given a hierarchy of classes ($r$, $a$,
		$b$, $c$, $d$, $e$, and $f$), the dashed boxes show the number of
		classifiers required for each method. Note that the \textit{lcn},
			\textit{lcpn}, \textit{lcl}, and global classifiers require 6, 4, 3,
			and 1 predictors, respectively.}
	\label{fig:hmc}
\end{figure}

Figure~\ref{fig:hmc} illustrates the four HMC methods used in this
work: \textit{Local classifier per node} (\textit{lcn}) consists of
training one binary classifier for each class in the hierarchy except
the root. \textit{Local classifier per parent node} (\textit{lcpn})
consists of training a multi-label classifier for each parent node in
the hierarchy to distinguish between its child classes. \textit{Local
	classifier per level} (\textit{lcl}) consists of training one
multi-label classifier for each level of the class hierarchy except
for the root. \textit{Global classifier} consists of building a single
multi-label classifier taking into account the hierarchy as a whole
during a single run. The global classifier can assign classes at
potentially every level of the hierarchy to an instance.

\subsection{SHAP feature contribution}

The performance of classification algorithms is partly determined by
the features used to train a particular predictor. SHAP (SHapley
Additive exPlanation) is a framework that computes the importance
values for each feature in a dataset using concepts from game
theory~\cite{lundberg-shap-2017,lundberg-shap-2020}. SHAP assigns
Shapely values to explain which features in the model are the most
important for prediction by calculating the changes in the prediction
when features are conditioned. Given a predictor and a training set,
SHAP computes a matrix with the same dimensions of the predictor's
output containing the Shapely values for each instance and class. For
example, in a binary classification problem and a training set of $n$
instances, the output of SHAP is a matrix of dimension $n\times2$
(there are two classes, positive and negative). In multi-label
classification problems, the output is a matrix of dimension $n\times
2$ for each class, since classes are not mutually exclusive and the
outcome is either positive or negative for each class.

\section{Clustering-based Feature Extraction}
\label{sec:method-feat}

The approach for extracting features from the GCN using a clustering
algorithm and Gene Ontology term enrichment is presented.  It combines
information from the GCN, and the associations between genes and
functions to create features capturing topological properties of the
GCN.

The inputs of the approach are a GCN, denoted by $G=(V,E,w)$, a set of
(biological) functions $A$, an annotation function $\phi:V\to2^A$, and
a set $K=\{k_0,\dots,k_{m-1}\}$ for sampling the number of clusters.
The annotation function $\phi$ must satisfy true-path rule for the GO
hierarchy~\cite{ashburner-go-2000,valentini-tpr-2009}. That is, if a
gene is associated to a function, then it must also be associated to
every ancestor of the function in the hierarchy, and if a gene is not
associated to a function, then it must not be associated to any of its
descendants.

The outputs are two feature matrices $J_G$ and $J_F$, of dimension
$V\times A\cdot K\rightarrow [0,1]$, specifying the likelihood of
the genes $V$ to be associated to the functions in $A$ when the graph
is decomposed in $m$ clusters. Matrices $J_G$ and $J_F$ correspond to
the GCN (that is the graph $G$) and an affinity graph defined the next
subsection.

\begin{figure}[ht!]
	\centering
	\includegraphics[width=\linewidth]{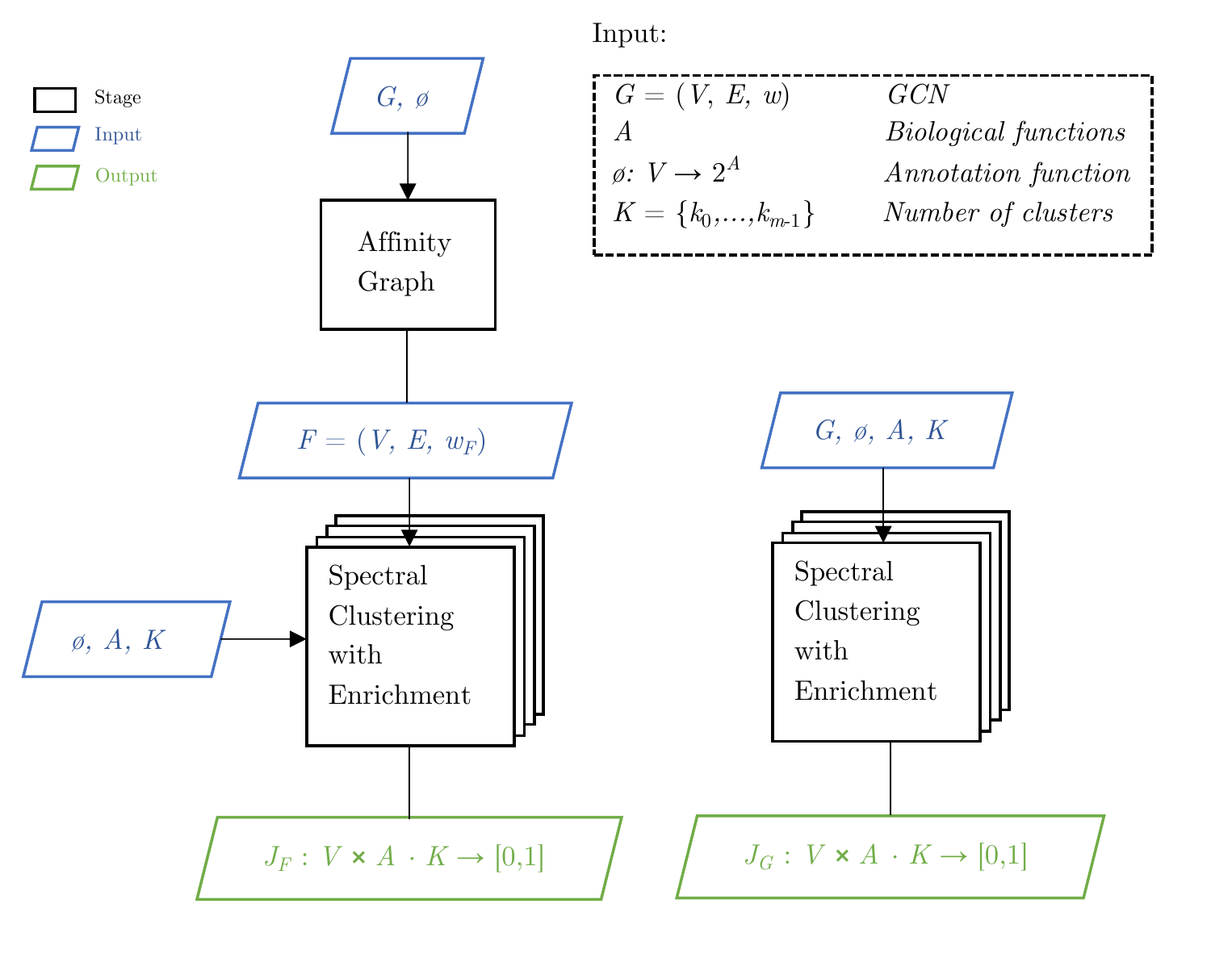}
	\caption{The clustering-based feature extraction approach consists of
	three stages. Namely, creation of affinity graph, clustering
	computation, and Gene Ontology term enrichment. Its inputs are a GCN,
	denoted by $G=(V,E,w)$, a set of functions $A$, an annotation
	function $\phi:V\to2^A$, and a set $K=\{k_0,\dots,k_{m-1}\}$.  Its
	output are two feature matrices (for both $G$ and its enriched
	version $F$) of dimension $V\times A\cdot K\rightarrow [0,1]$ that
	specify how likely it is for the genes to be associated to the
	functions in $A$ when the graph is decomposed $m$ clusters, each of
	size $k_i$, for $0\leq i\leq m$.}
	\label{fig:wflow-feat}
\end{figure}

The feature extraction approach consists of three stages, which are
depicted in Figure~\ref{fig:wflow-feat}. First, an affinity graph $F$
with information in $\phi$ is created from $G$. Second, the spectral
clustering algorithm is applied to both $G$ and its enriched version
$F$ for the $m$ different number of clusters specified in $K$. Third,
the Gene Ontology term enrichment technique is used to create $m$
features for each function $a\in A$, corresponding to the number of
clusters in $K$.

\subsection{Affinity Graph Creation}

An affinity graph $F = (V, E, w_F)$ between $G$ and $\phi$ is built.
Its weight function is defined as the mean between the co-expression
weight specified by $w$ and the proportion of shared functions between
genes specified by $\phi$.

\begin{definition}\label{def.affinity}
	The \emph{weight function} $w_F : V\times V \to [0,1]$ is defined for any
	$u,v \in V$ as
	\begin{align*}
	w_F(u,v) = \frac{1}{2}\left(\frac{w(u,v) -1}{\max(w)-1}+\frac{|\phi(u) \cup \phi(v)|}{|\phi(u) \cap \phi(v)|}\right),
	\end{align*}
	where $\max(w)$ denotes the maximum value in the range of $w$ (which
	exists because $w$ is finite).
\end{definition}

\noindent
Under the assumption that at least one element in the range of $w$ is
greater than 1, it is guaranteed that the range of $w_F$ is $[0,1]$
(because $w: V \times V \to [1,\infty)$). This is indeed the case, in
practice, because the co-expression between two genes in the GCN is
quantified in terms of the $z$-score, which is highly unlikely to be
1 for all pairs of genes.

\subsection{Gene Clustering}

The spectral clustering algorithm is applied independently to each
graph $X \in \{G, F\}$ to decompose $X$ (i.e., group the genes $V$)
using the number of clusters specified by $K=\{k_0,\dots,k_{m-1}\}$.
The decomposition of $X$ is performed $m$ times, once per $k$ in $K$.
The adjacency matrices of the weighted and undirected graphs $G$ and
$F$ are used as the precomputed affinity matrices required for the
spectral clustering algorithm. The outcome of the clustering algorithm
is an assignment from nodes to clusters of size $k$, for each $k\in
K$. More precisely, the outputs of this stage are the matrices $I_X :
V \times K \to [0,1]$, where each column $0\leq i < m$ represents the
decomposition of $X$ in $k_i$ clusters.

\subsection{Gene Enrichment}

The goal of this stage is to produce a matrix $J_X : V \times A \cdot
K \to [0,1]$ for each $X \in \{G, F\}$, specifying how likely it is
for the genes to be associated to every function $a\in A$ when $X$ is
decomposed in the given number of clusters.

For each decomposition from the previous stage (i.e., each column of
the matrices $I_X$) and function $a\in A$, the resulting clusters are
used to compute whether a significant number of members associated to
function $a$ is (locally) present. Intuitively, if genes that are grouped
together have a strong co-expression relation and most of the group
are associated to gene function $a$, then the remaining genes are also
likely to be associated to $a$ (i.e., guilt by association,
see~\cite{petsko-gba-2009}). In this way, for each $v \in V$, $a\in
A$, and $k \in K$, the entry $J_X(v, a\cdot k)$ is a $p$-value
indicating if the function $a$ is over-represented in the
decomposition of $k$ clusters of $X$. This process is commonly known
as Gene Ontology term enrichment and may use different statistical
tests, such as, Fisher's exact test~\cite{yonrhee-goenr-2008}.

\section{Hierarchical Multi-label Classification for Gene Function Prediction}
\label{sec:method-pred}

This section presents the approach for gene function prediction using
HMC to create a predictor, enriched with the information of the
features created in Section~\ref{sec:method-feat}.

The GO hierarchy is defined as a directed acyclic graph (DAG)
containing three main types of annotations: biological processes,
molecular functions, and cellular component~\cite{go-go-2019}. This
work focuses on biological processes, i.e., a subgraph of the GO
hierarchy that contains 28 roots (i.e., functions in the GO hierarchy
with null indegree). This subgraph is denoted as $H=(A,R)$, where $A$
is the set of biological processes and $R$ the binary relation
representing ancestral relations between pairs of biological processes
(i.e., $(a,b)\in R$ means that function $b$ is ancestor of function
$a$ in the GO hierarchy). The topological-sorting traversal algorithm
presented in~\cite{romero-hmc-2022} is used to transform the GO
hierarchy of biological processes into a tree. As a result, the
hierarchy is split into several components, i.e., subtrees of $H$
called sub-hierarchies. Each sub-hierarchy, $H'=(A',R')$ with
$A'\subseteq A$, $R'\subseteq R$, and $r\in A'$ the root, is
associated to a subgraph $G'=(V',E',w)$ containing all genes $v\in V$
associated to $r$, i.e., $V'=\phi^{-1}(r)$. Note that, the proposed
approach is independently applied to each sub-hierarchy.

The inputs of the approach are a sub-hierarchy $H'=(A',R')$, a
subgraph of the GCN, denoted by $G'=(V',E',w)$, where $V'\subseteq V$
and $E'\subseteq E$, an annotation function $\phi:V\to2^{A'}$, the
matrices $J_G$ and $J_F$ resulting from Section~\ref{sec:method-feat},
and a constant value $c\in[0,1]$ for feature selection. The output is
a function $\psi : V' \times A'\to [0,1]$, specifying, for each gene
$v \in V'$, the probability $\psi(v,a)$ of $v$ being associated to
function $a\in A'$.

\begin{figure}[ht!]
	\centering
	\includegraphics[width=\linewidth]{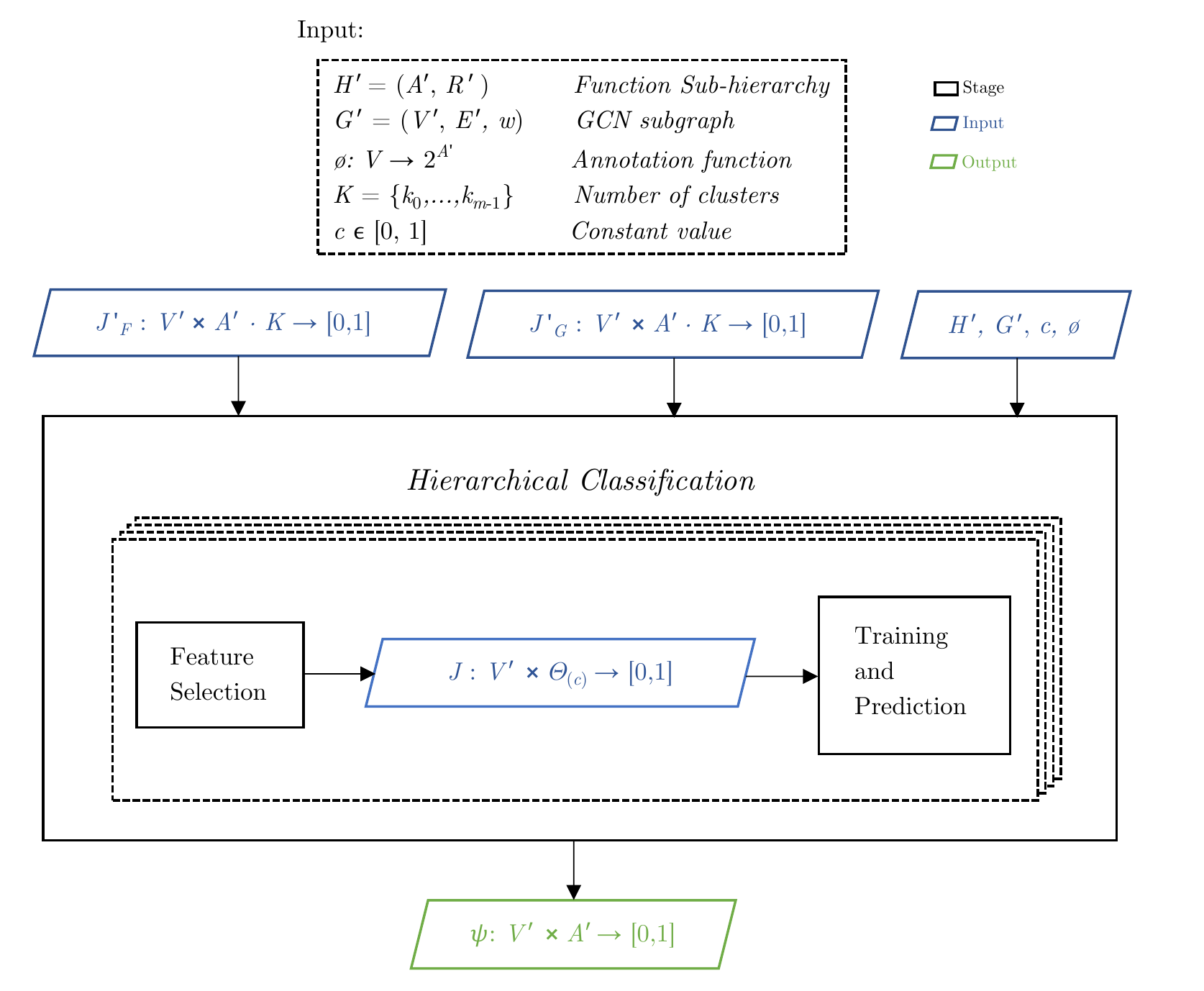}
	\caption{The prediction approach mainly consists of two stages,
		feature selection with SHAP and hierarchical multi-label
		classification. Its inputs are a sub-hierarchy $H'=(A',R')$, a
		subgraph of the GCN $G'=(V',E',w)$, an annotation function
		$\phi:V\to2^{A'}$ that satisfy the sub-hierarchy $H'$, the
		sub-matrices of $J_G$ and $J_F$ containing only the functions $A'$ and
		genes $V'$, and a constant value $c\in[0,1]$ for feature selection.
		Its output is a function $\psi : V' \times A'\to [0,1]$, which
		indicates for each gene $v \in V'$, the probabilities $\psi(v,a)$ of
		$v$ being associated to function $a\in A'$.}
	\label{fig:wflow-pred}
\end{figure}

First, sub-matrices $J_G'$ and $J_F'$ are created from $J_G$ and
$J_F$, by respectively considering only the genes $V'\subseteq V$ and
functions $A'\subseteq A$. These sub-matrices represent structural
properties of the GCN subgraph $G'$, and associations between genes
and functions based on multiple partitions of each graph.
Figure~\ref{fig:wflow-pred} illustrates the prediction approach. The
reminder of this section is devoted to detailing the prediction
approach.

\begin{figure}
	\centering
	\includegraphics[width=0.45\linewidth]{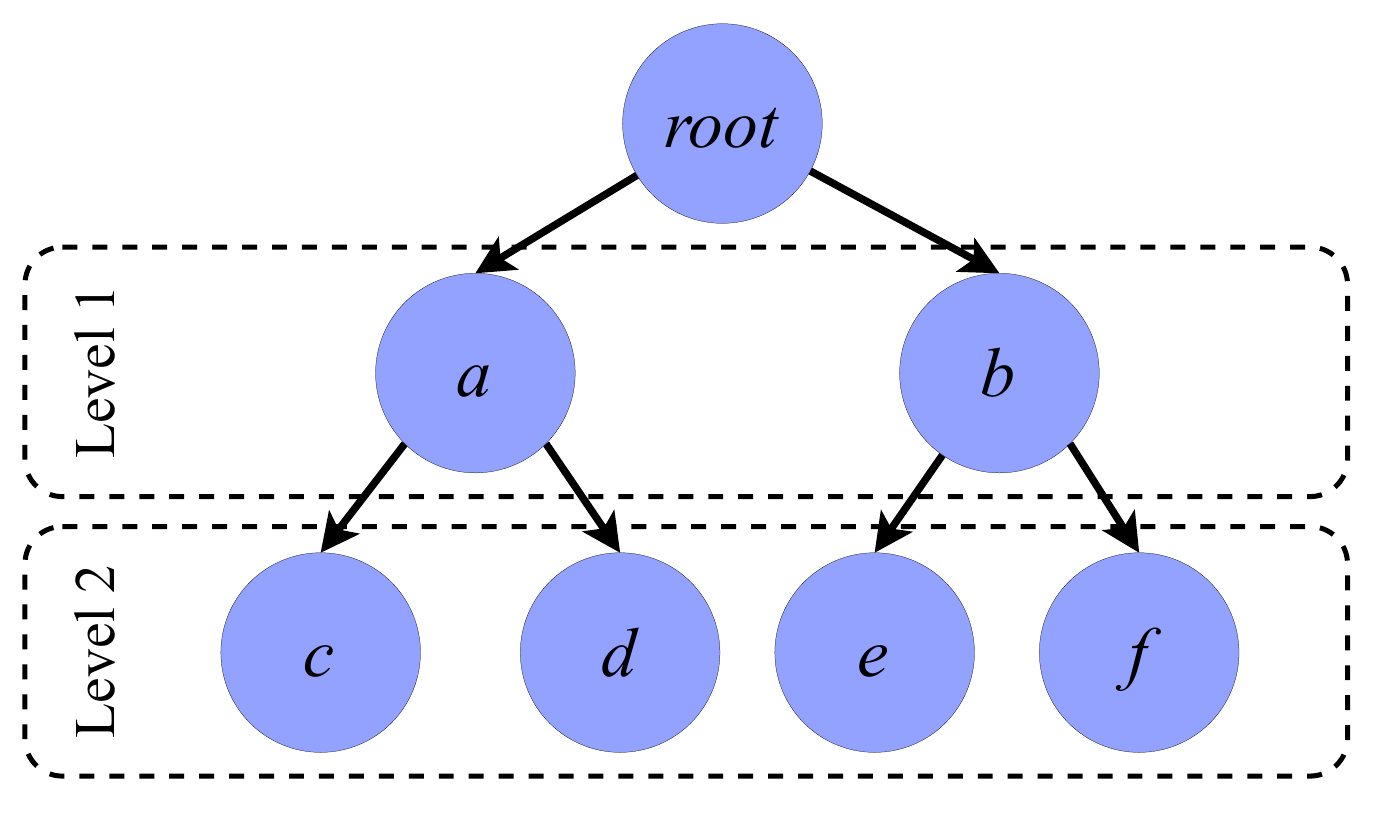}
	\caption{Gene function prediction considering the function hierarchy
	and using a local classifier per level method. The predictor for the
	level~1 predicts functions $a$ and $b$, so only the features from
	$J_G'$ and $J_F'$ associated to functions $a$ and $b$ are considered
	for the feature selection.}
	\label{fig:hmc-fs}
\end{figure}

SHAP filters the extracted features with more impact in the prediction
task, and HMC is used to predict associations between genes and
functions without inconsistencies (i.e., complying the true-path
rule). Since local HMC methods use more than one predictor per
hierarchy, the feature selection is executed for each predictor
independently, considering only the features related to the functions
being predicted, denoted by $A''\subseteq A'$. For example, consider
the function hierarchy and a local classifier per level method
depicted in Figure~\ref{fig:hmc-fs}. The predictor for level~1
predicts functions $a$ and $b$, so only the features associated to
functions $a$ and $b$ are considered for the feature selection.

\subsection{Feature Selection}

The aim of feature selection is to produce a matrix $J : V' \times
\Theta(c) \to [0, 1]$ by selecting a reduced number of significant
features from $J_G'$ and $J_F'$. The number of selected features is
denoted by $0 \leq \Theta(c)\leq 2m\cdot|A''|$, where $m\cdot|A''|$ is
the number of features in each matrix $J_G'$ and $J_F'$, denoted as
$q$ (that is $q=m\cdot|A''|$).

Feature selection is conveyed from $J_G'$ and $J_F'$ to $J$ using
SHAP. Let $J_{G+F}'$ denote the matrix resulting from extending $J_G'$
with the $q$ features of $J_F'$. That is, for each $v \in V'$, the
expression $J_{G+F}'(v, \_)$ denotes a function with domain $[0, 2q)$
and range $[0,1]$, where the values in $[0, q)$ denote the $p$-values
associated to $v$ in $G$ and the values in $[q, 2q)$ the ones
associated to $v$ in the enriched version of $G$. For each entry
$J_{G+F}'(v, j)$, with $v \in V'$ and $0 \leq j < 2q$, the mean
absolute SHAP value $s_{(v,j)}$ is computed after a large enough
number of Shapely values are computed (executions of SHAP). Features
are selected based on the cutoff
\[c \cdot \sum_{j=0}^{2q-1}s_{(v,j)},\]
i.e., on the sum of mean absolute values by a factor of the input
constant $c$. The first $\Theta(c)$ features, sorted from greater to
lower mean absolute SHAP value, are selected as to reach the given
cutoff.

Note that the input constant $c$ is key for selecting the number of
significant features. The idea is to set $c$ so as to find a balance
between prediction efficiency and the computational cost of building
the predictor.

\subsection{Training and Prediction}

This stage comprises a process that combines two supervised machine
learning techniques/tools to build the predictor $\psi$. In
particular, stratified $k$-fold cross-validation and hierarchical
multi-label classification are used sequentially in a pipeline.

The pipeline takes as input the matrix $J$, which specifies the
significant features of $J_G'$ and $J_F'$, the sub-hierarchy $H'$ and
the annotation function $\phi$.
First, $k$-fold is applied to split the dataset into $k$ different
folds for cross validation (note that $k$ is not related to the input
$K$). That is, each fold is used as a test set, while the remaining
$k-1$ folds are used for training. Recall that $k$-fold cross
validation aims to overcome overfitting in training. Furthermore, one
or multiple random forest classifiers are build and used for
prediction, the number of classifiers depends on the HMC method. 
Randoms forest is selected for this approach since it is a tree-based
and multi-label classification algorithm, which is interpretable (SHAP
can be applied). The parameter values used for random forest
classifiers, differently from the default scikit-learn values, are:
200 estimators (\textit{n\_estimators}) and minimum number of samples
of 5 (\textit{min\_samples\_split}).

Additionally, some HMC methods require an extra step to keep
prediction consistent w.r.t. the sub-hierarchy $H'$ (i.e., comply the
true-path rule). The probability of association between a function
$v\in V'$ and a function $a\in A'$ must be lower than the probability
of association between the same gene and the ancestor of $a$ in $H'$.
To satisfy this constraint cumulative probabilities are computed
throughout the paths in $H'$. That is, for each gene $v\in V$ and
functions $(a,b)\in R$, the predicted probability of the association
between $v$ and $a$ is multiplied by the predicted probability of
association between $v$ and $b$ (its ancestor). This process is
repeated for every path in the hierarchy from the root to the leaves.

The output of this stage is the predictor $\psi$, i.e., the
probabilities of associations between the genes in $V'$ and functions
$A'$. Note that the predictor $\psi$ satisfies the true-path rule.

\subsection{Performance Evaluation}

It is often the case in HMC datasets that individual classes have few
positive instances. In genome annotation, typically only a few genes
are associated to specific functions. This implies that for most
classes (deeper in the hierarchy), the number of negative instances by
far exceeds the number of positive instances. Hence, the real focus is
recognizing the positive instances (predict associations between genes
and functions), rather than correctly predicting the negative ones
(predict that a function is not associated to a given gene). Although
ROC curves are better known, their area under the curve is higher if a
model correctly predicts negative instances, which is not suitable for
HMC problems.

For this reasons, the measures (based on the precision-recall (PR) curve)
introduced by~\cite{vens-hmc-2008} are used for evaluation.

\paragraph{Area under the average PR curve.}
The first metric transforms the multi-label problem into a binary one
by computing the precision and recall for all functions $A'$ together.
This corresponds to micro-averaging the precision and recall.

The output of the prediction stage are the probabilities of
associations between genes $V'$ and functions $A'$. Thereby,
instead of selecting a single threshold to compute precision and
recall, multiple thresholds are used to create a PR
curve. In the PR curve each point represent the precision and recall
for a give threshold that can be computed as:

\begin{equation*}
\overline{\textnormal{Prec}} = \frac{\sum_{i}TP_i}{\sum_{i}TP_i+\sum_{i}FP_i}, \quad \textnormal{and} \quad
\overline{\textnormal{Rec}}=\frac{\sum_{i}TP_i}{\sum_{i}TP_i+\sum_{i}FN_i}.
\end{equation*}

Note that $i$ ranges over all functions $A'$, i.e., precision and
recall are computed for all functions together. The area under this
curve is denoted as AU($\overline{\textnormal{PRC}}$).

\paragraph{Average area under the PR curves.}
The second metric corresponds to the (weighted) average of the areas
under the PR curves for all functions $A'$. This metric, referred as
macro-average of precision and recall, can be computed as follows:

\begin{equation*}
\overline{\textnormal{AUPRC}}_{w_1,w_2,\dots,w_{|A'|}} = \sum_{i}w_i\cdot \textnormal{AUPRC}_i.
\end{equation*}

If the weights of all functions are the same (i.e., $1/|A'|$) the
metric is denoted as $\overline{\textnormal{AUPRC}}$. In addition,
weights can also be defined based on the number of genes associated to
functions in $\phi$, i.e., $w_a=|\phi^{-1}(a)|/\sum_i |\phi^{-1}(i)|$
for $a\in A$. In the later case, denoted as
$\overline{\textnormal{AUPRC}_w}$, more frequent functions get higher
weight. Note that one point in the weighted PR curve corresponds to
the (weighted) average of the AUPRC of all functions $A'$ given a
threshold.

\section{Case study: \textit{Zea mays}}
\label{sec:case}

Next section describes a case study on applying the feature extraction
and prediction approach presented in Sections~\ref{sec:method-feat}
and~\ref{sec:method-pred} to maize (\textit{Zea mays}). First, the
maize data used for the case study is described. Second, the proposed
approach is applied to the maize data. Lastly, the performance of the
proposed approach is compared to two models trained using each set of
features $J_G$ and $J_F$, independently.

\subsection{Data Description and Feature Extraction}

The co-expression information used in the study is imported from the
ATTED-II da\-ta\-base~\cite{obayashi-atted2018-2018}. The gene
co-expression network $G = (V, E, w)$ comprises \numprint{26131}
vertices (genes) and \numprint{44621533} edges. In this case, a
$z$-score threshold of $1$ is used as the cut-off measure for $G$,
i.e., $E$ contains edges $e$ that satisfy $w(e) \geq 1$ (most of them
satisfying $w(e) >1$). Note that the highest value is assigned to the
strongest connections. The functional information for this network is
taken from DAVID Bioinformatics Resources~\cite{huang-david-2009}
(2021 update); it contains annotations of biological processes, i.e.,
pathways to which a gene contributes. It is important to note that
genes may be associated to several biological processes, and
biological processes may be associated to multiple genes. The database
comprises \numprint{3924} biological processes $A$ and \numprint{7021}
ancestral relations $R$ between these functions, that represent the
hierarchy $H=(A,R)$ of the GO~\cite{go-go-2019}. A total of
\numprint{255865} association between genes and functions are
considered, these associations represent the annotation function
$\phi:V\rightarrow2^A$.

The feature extraction approach is applied with the inputs $G$, $A$,
$\phi$ and $K=\{10,20,\dots,100\}$ (values are incremented in steps of
10 up to 100). The outputs are the feature matrices $J_G$ and $J_F$
that specify how likely it is for the maize genes $V$ to be associated
to the biological processes $A$ when the graph is decomposed in the
number of clusters in $K$.

Moreover, only functions associated to more than 200 genes have been
considered, so the number of functions in the resulting
sub-hierarchies is tractable regarding the dimension of the output of
SHAP (see Section~\ref{sec:prelim}). Recall that the Gene Ontology
hierarchy splits into 28 sub-hierarchies when considering only
biological processes. Additionally, all sub-hierarchies with less than
10 functions are discarded and the topological-sorting algorithm
introduced in~\cite{romero-hmc-2022} is used to transform the
sub-hierarchies, represented as DAGs, into trees. For each ancestral
relation $(a,b)\in R$ ($b$ is ancestor of $a$), the algorithm assigns
a weight as the ratio of the number of genes associated to the $a$ to
the number of genes associated to $b$. Then, for each function $a\in
A'$ with more than one parent, only the one with the higher weight
remains (ties are broken arbitrarily).

\begin{table}[htbp!]
	\centering
	\begin{tabular}{|p{1.8cm}|p{2.8cm}|r|r|p{2.8cm}|}
		\hline
		Root &                   Description & Functions & Genes & Functions per level\\
		\hline\hline
		GO:0050896 &   response to stimulus &    13 &  1733 & 5/5/2\\
		GO:0051179 &           localization &    25 &  1497 & 3/5/9/6/1\\
		GO:0065007 &  biological regulation &    37 &  2647 & 2/5/11/10/4/2/2\\
		GO:0008152 &      metabolic process &    92 &  6596 & 8/18/38/12/7/6/2\\
		GO:0009987 &       cellular process &    92 &  8005 & 13/19/19/17/13/8/2\\
		\hline
	\end{tabular}
	\caption{Resulting sub-hierarchies $H'$ of biological processes for
		maize. The identifier and description of each root function $r$ is
		presented in the first and second columns, respectively. The third
		column shows the number of functions $A'$ within each sub-hierarchy
		and the fourth column shows the number maize genes in the GCN
		subgraph $G'$ associated to $H'$. The last column shows the number of
		functions per level, e.g., the first sub-hierarchy has 3 levels and
		there are 5, 5, and 2 functions on each level.}
	\label{tab:func}
\end{table}

\begin{figure}[htbp!]
	\centering  
	\includegraphics[width=0.7\linewidth]{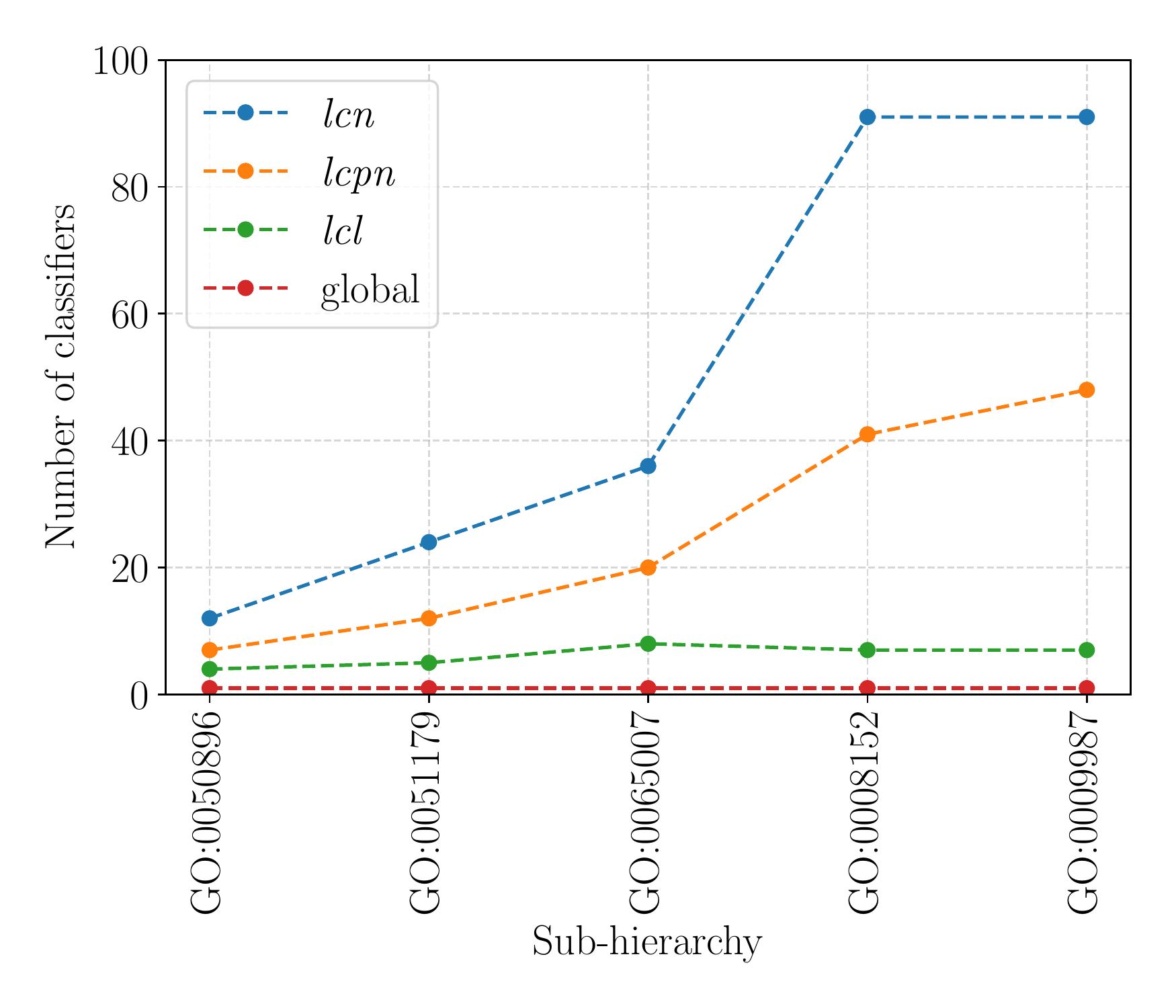}
	\caption{number of classifiers trained per HMC method and
	sub-hierarchy. The \textit{lcn} requires $|A'|-1$ classifiers. The
	\textit{lcpn} requires as many classifiers as functions with children
	in $H'$. The \textit{lcl} requires as many classifiers as the number
	of levels in $H'$. At last, the global method requires one classifier
	per hierarchy.}
	\label{fig:hmc-class}
\end{figure}

As result, there are 5 sub-hierarchies of biological processes.
Table~\ref{tab:func} describes each sub-hierarchy $H'$, starting by
the root term  $r$ and its description, following the number of
functions $A'$ and the number of genes $V'$ in the associated GCN
subgraph $G'$. The prediction approach is applied to each
sub-hierarchy $H'$ independently. The remaining input parameter for
the prediction approach is $c=0.9$ (recall that this parameter is used
to filter the most relevant features according to their mean SHAP
value). Figure~\ref{fig:hmc-class} depicts the number of classifiers
trained per HMC method and sub-hierarchy. Note that the global method
requires one classifier per hierarchy, while the \textit{lcn} requires
$|A'|-1$ classifiers.

\subsection{Summary of Results}

\begin{figure}[ht!]
	\centering  
	\includegraphics[width=0.8\linewidth]{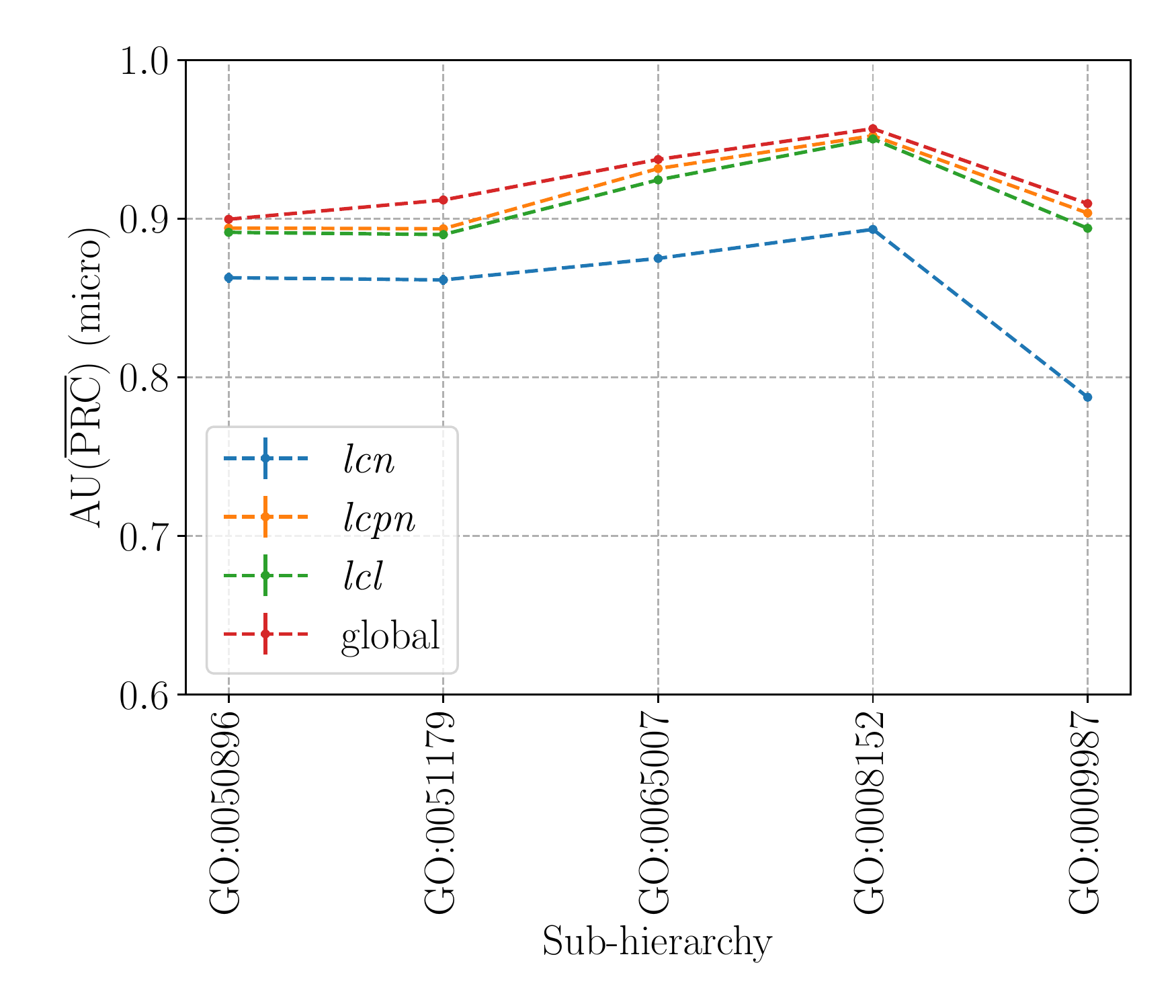}
	\caption{Prediction performance of the proposed approach measured
		with the area under the average PR curve, i.e.,
		AU($\overline{\textnormal{PRC}}$). The performance is measured
		independently per sub-hierarchy.}
	\label{fig:micro}
\end{figure}

Figure~\ref{fig:micro} presents the prediction performance of the
proposed approach measured with the AU($\overline{\textnormal{PRC}}$)
(denoted as \textit{micro}) for four HMC methods, namely, local
classifier per node (\textit{lcn}), local classifier per parent node
(\textit{lcpn}), local classifier per level (\textit{lcl}), and global
classifier. In general, it can be seen that all methods get a high
area under the average PR curve, but the global classifier outperforms
the local methods for all sub-hierarchies. The proposed approach
identifies the associations between genes and functions by using the
features extracted from the GCN $G$ and the affinity graph $F$, and
considering the ancestral relations of the biological processes. The
global method obtains the best performance, followed by the
\textit{lcpn} and the \textit{lcl}. Using multi-label classifiers is
better than using a binary classifier for each function, i.e.,
\textit{lcn} method.

\begin{figure}[htbp!]
	\centering  
	\includegraphics[width=0.8\linewidth]{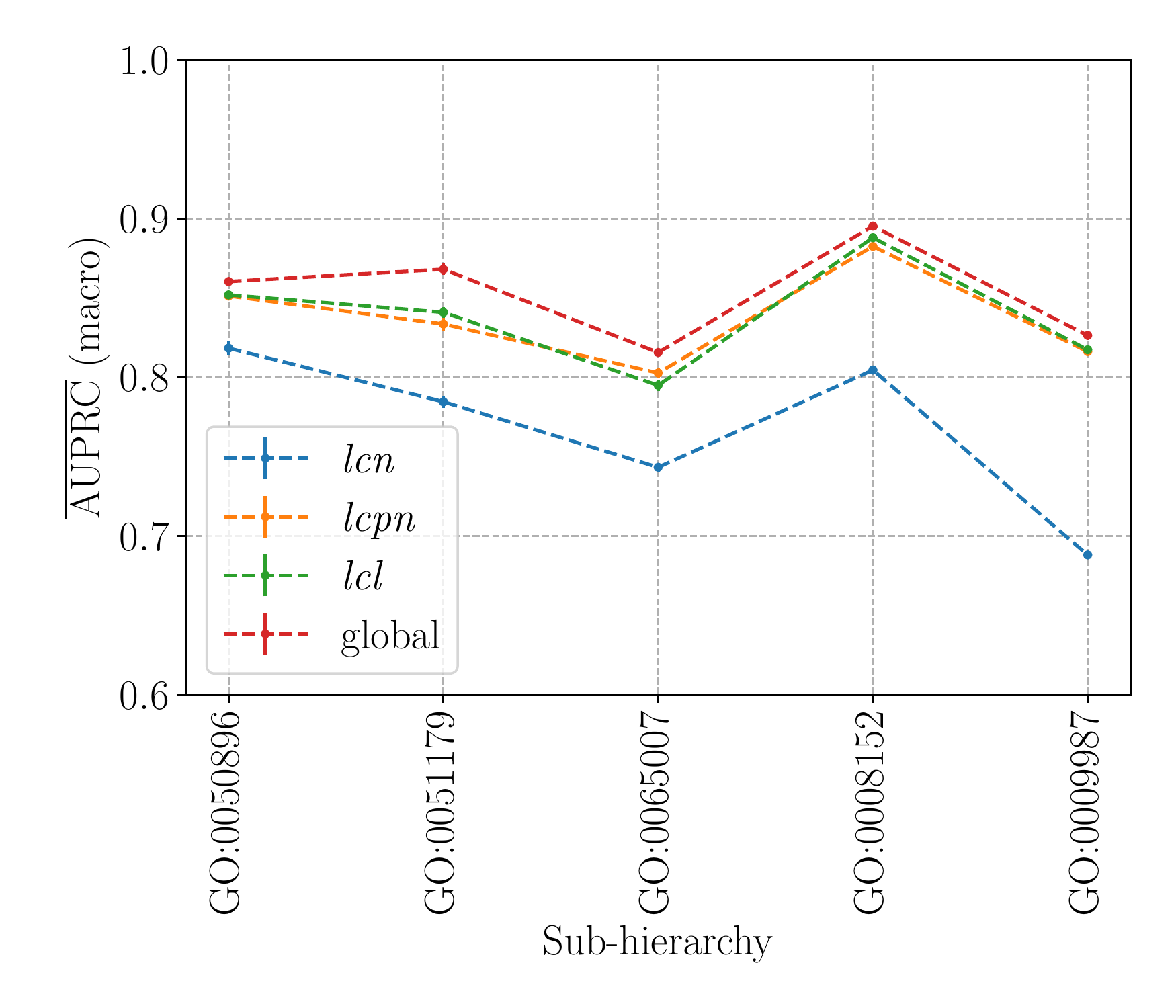}
	\caption{Prediction performance of the proposed approach measured
		with the average area under the PR curve, i.e.,
		$\overline{\textnormal{AUPRC}}$. The performance is measured
		independently per sub-hierarchy.}
	\label{fig:macro}
\end{figure}

The micro score measures the overall performance of all functions
within a sub-hierarchy without distinguishing between them.
Figure~\ref{fig:macro} presents the prediction performance measured
with the $\overline{\textnormal{AUPRC}}$, denoted as \textit{macro}.
The macro score measure the prediction performance for each function
individually and then takes the average. The conclusion is similar,
the global method outperforms the local ones. 

\begin{figure}[htbp!]
	\centering  
	\includegraphics[width=0.8\linewidth]{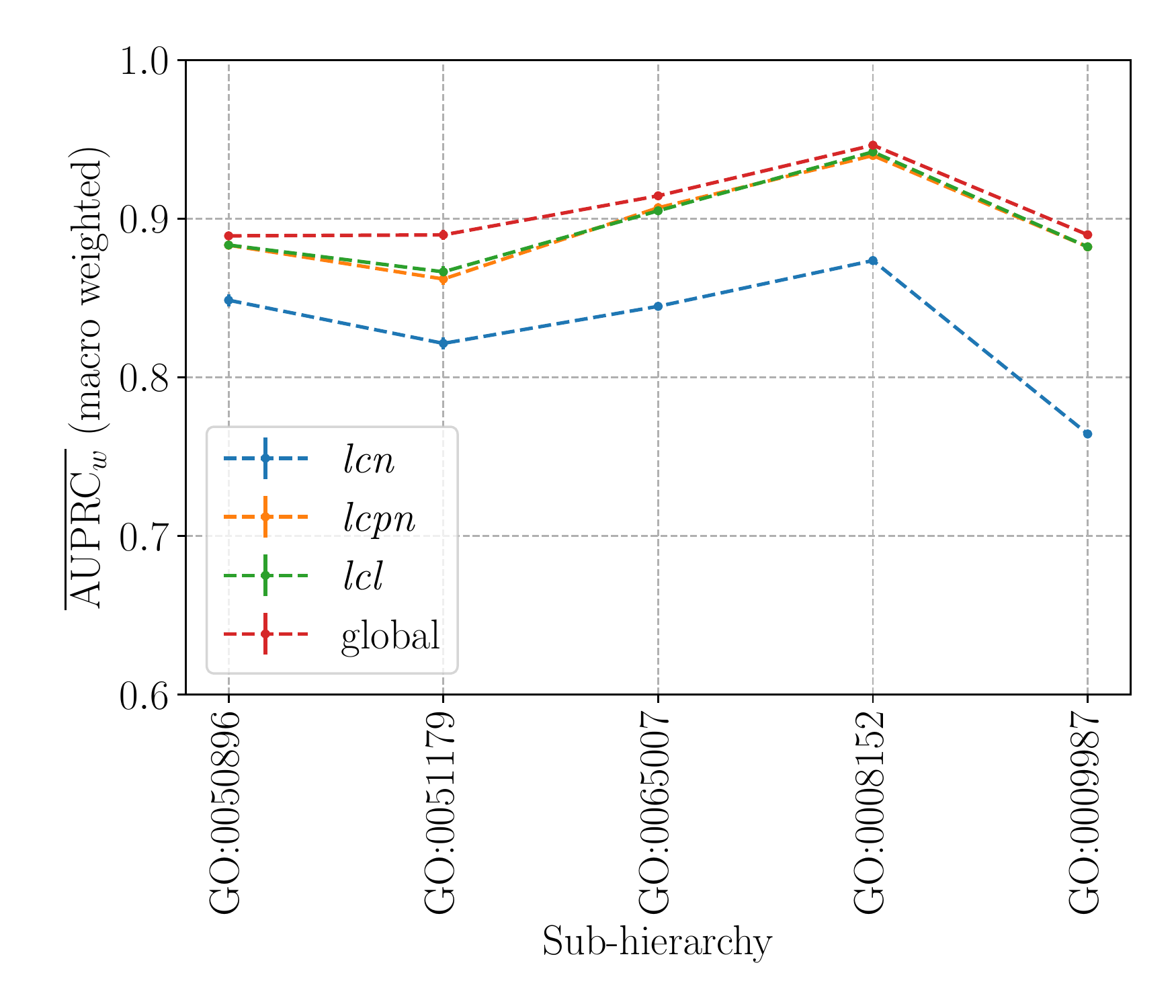}
	\caption{Prediction performance of the proposed approach measured
		with the average area under the PR curve, i.e.,
		$\overline{\textnormal{AUPRC}_w}$. The performance is measured
		independently per sub-hierarchy.}
	\label{fig:macrow}
\end{figure}

Finally, Figure~\ref{fig:macrow} illustrates the prediction
performance measured with the $\overline{\textnormal{AUPRC}_w}$,
denoted as \textit{macro weighted}. This score weights the individual
performance of each function according to the number of genes
associated to it. Thereby, the leaves and deeper functions in a
sub-hierarchy always get lower weight than the others. Note that the
deeper a functions is in a sub-hierarchy, the lower the predicted
probabilities becomes. The global method outperforms the locals again.
The conclusion is consistent with the three metrics, using clustering
techniques to extract features from the GCN and considering the
hierarchical structure of the biological processes seems to be key for
the gene function production task.

\begin{table}
	\centering
	\begin{tabular}{|l|r|r|}
		\hline
		Root & Total & Filtered \\ \hline\hline
		GO:0050896 &  239 & 124 \\
		GO:0051179 &  479 & 263 \\
		GO:0065007 &  713 & 402 \\
		GO:0008152 & 1812 & 796 \\
		GO:0009987 & 1813 & 853 \\\hline
	\end{tabular}
	\caption{Number of extracted and filtered features used for the
		global method per sub-hierarchy. Recall that the extracted features
		are filtered using the mean SHAP values to select the more important
		with a cutoff defined by the input constant $c$.}
	\label{tab:feats}
\end{table}

It has been shown in~\cite{romero-clust-2022} that the new features
built from the GCN, and the associations between genes and functions
with the spectral clustering algorithm are key to improve the
prediction performance in the gene annotation problem (w.r.t. other
features of the GCN and gene functional information). However, the
feature extraction approach presented in Section~\ref{sec:method-feat}
produces two different sets of features, namely, $J_G$ and $J_F$, that
are combined and used for prediction. The individual relevance of each
set of features for the gene annotation problem is analyzed by (i)
looking at the distribution of the filtered features for the global
method and (ii) comparing the performance of the prediction task using
each set of features independently. Table~\ref{tab:feats} presents the
number of extracted and filtered features used for the global method
per sub-hierarchy. Recall that the features are filtered using the
mean SHAP values to select the more important ones with a cutoff
defined by the input constant $c$.

\begin{figure}[htbp!]
	\centering  
	\includegraphics[width=0.7\linewidth]{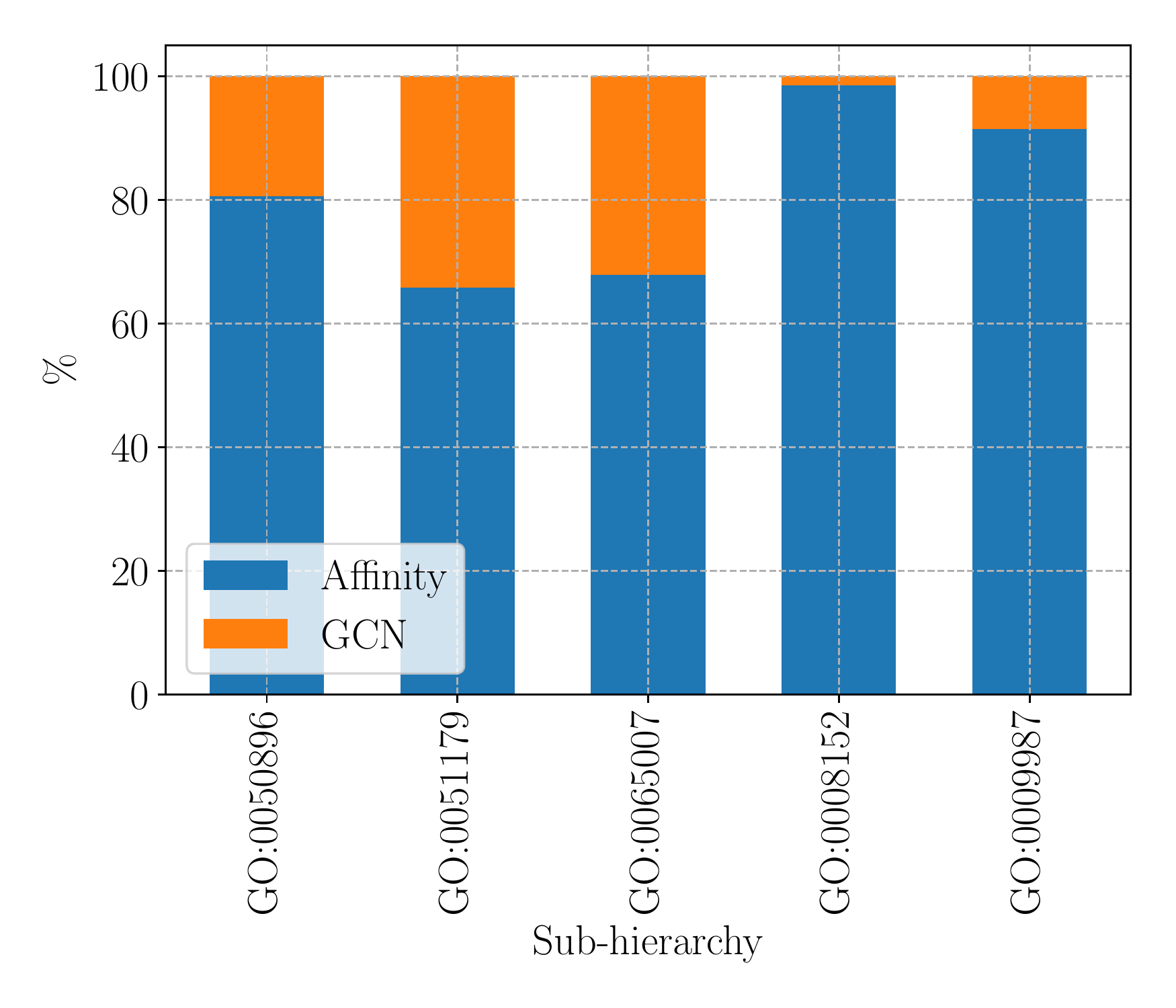}
	\caption{Distribution of the filtered features from $J_G$ and $J_F$
		for the global method per sub-hierarchy.}
	\label{fig:feats}
\end{figure}

\begin{figure}[htbp!]
	\centering  
	\includegraphics[width=0.75\linewidth]{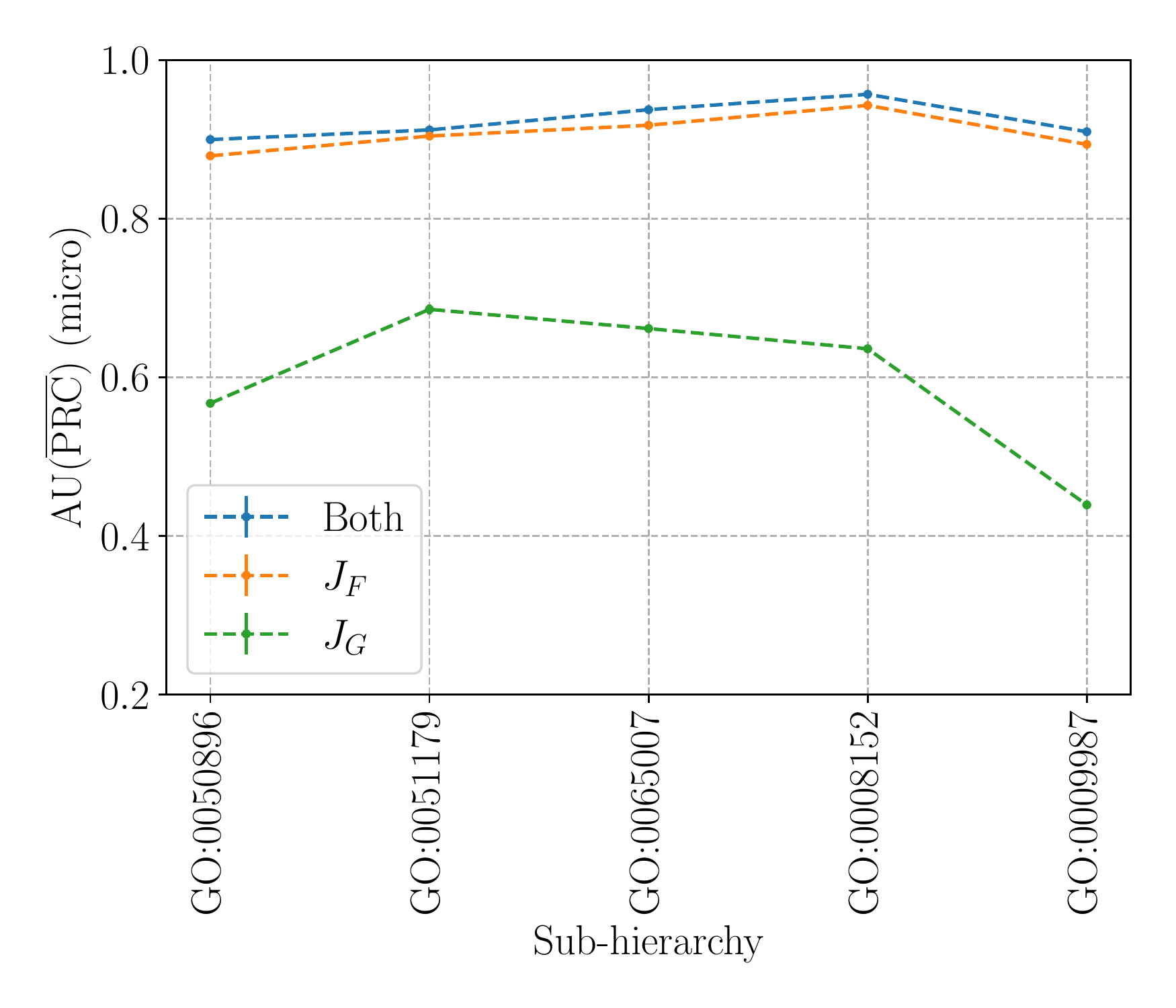}
	\includegraphics[width=0.75\linewidth]{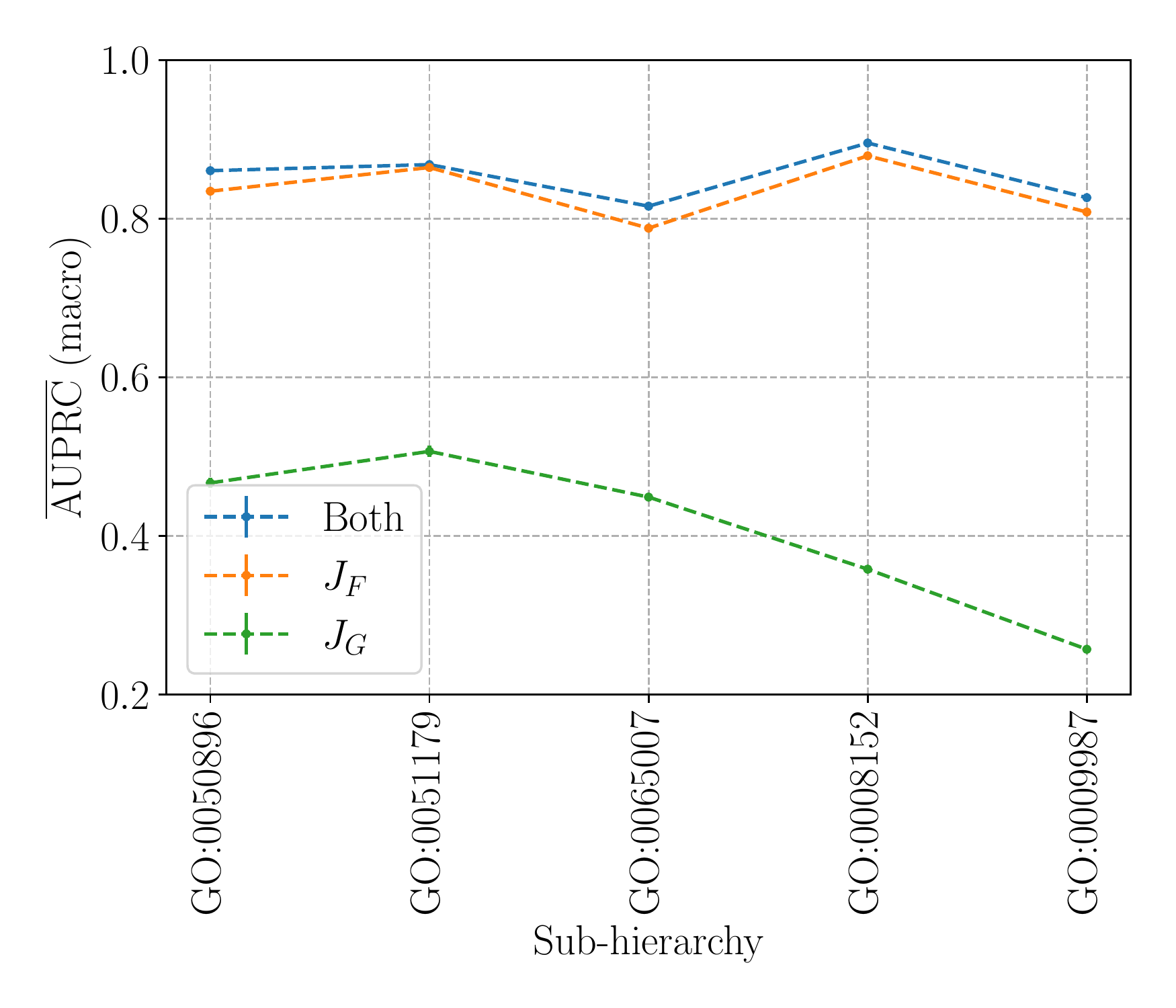}
	\caption{Prediction performance of the global method trained using
		the features $J_G$ and $J_F$ independently, and the proposed approach
		(i.e., their combination) measured with
		AU($\overline{\textnormal{PRC}}$) and
		$\overline{\textnormal{AUPRC}}$. The performance is measured
		independently per sub-hierarchy.}
	\label{fig:compfeat}
\end{figure}

Figure~\ref{fig:feats} illustrates the distribution of the filtered
features for the global method per sub-hierarchy. Note that, even
though the features from the affinity graph $F$ (i.e., $J_F$) are more
important, features from the GCN $G$ (i.e., $J_G$) are also selected
for all sub-hierarchies. Figure~\ref{fig:compfeat} shows the
prediction performance of the global HMC method trained using the
features $J_G$ and $J_F$ independently, and the proposed approach
(i.e., their combination) measured with
AU($\overline{\textnormal{PRC}}$) and $\overline{\textnormal{AUPRC}}$.
The combination of both sets of features, extracted from the GCN and
the affinity graph is key to improve the performance of the proposed
approach for all sub-hierarchies.

\section{Related Work and Concluding Remarks}
\label{sec:concl}

\subsection{Related Work} 

Zhou et al.~\cite{zhou-gcn-2020} presented an approach to predict
functions of maize proteins using graph convolutional networks. In
particular, an amino acid sequence of proteins and the GO hierarchy
were used to predict functions of proteins with a deep graph
convolutional network model (DeepGOA). Their results showed that
DeepGOA is a powerful tool to integrate amino acid data and the GO
structure to accurately annotate proteins.
Similarly, the work presented in~\cite{cruz-enrich-2020} aims to
predict the phenotypes and functions associated to maize genes using:
(i) hierarchical clustering based on datasets of transcriptome (set of
molecules produced in transcription) and metabolome (set of
metabolites found within an organism); and (ii) GO enrichment
analyses. Their results showed that profiling individual plants is a
promising experimental design for narrowing down the lab-field gap.
Gligorijevi{\'c} et al.~\cite{gligorijevic-deepnf-2018} proposed a
network fusion method based on multimodal deep autoencoders to extract
high-level features of proteins from multiple interaction networks.
This method, called \textit{deepNF}, relied on a deep learning
technique that captures relevant protein features from different
complex, non-linear interaction networks. Their results showed that
extracting new features from biological networks is key to annotate
gene with functions.
The work in~\cite{zhao-hphash-2019} is also closely related. They
presented Gene Ontology hierarchy preserving hashing (HPHash), a gene
function prediction method that retains the hierarchical order between
GO terms. It used a hierarchy preserving hashing technique based on
the taxonomic similarity between terms to capture the GO hierarchy.
Hashing functions were used to compress the gene-term association
matrix, where the semantic similarity between genes was used to
predict the functions of the genes. Their results showed that HPHash
preserves the GO hierarchy and improves prediction performance.

In addition, the authors in~\cite{chen-ifeature-2018} presented
\textit{iFeature}, a Python-based toolkit for generating numerical
feature representation schemes from protein sequences. It integrated
algorithms for feature clustering, selection, and dimensionality
reduction to facilitate training, analysis, and benchmarking of
machine-learning models.
In a related way, Mu et al.~\cite{mu-fegs-2021} showed that feature
extraction of protein sequences is helpful for prediction of protein
functions or interactions. They introduced FEGS (Feature Extraction
based on Graphical and Statistical features), a novel feature
extraction model for protein sequences that combines graphical and
statistical features. Their results showed that similarity analysis of
protein sequences has applications in the study of gene annotation,
gene function prediction, identification and construction of gene
families, and gene discovery.

\subsection{Concluding Remarks and Future Work} 

By combining network-based modeling, cluster analysis, interpretable
machine learning, and hierarchical multi-label classification, the
approach presented in this paper introduces a novel method to address
the gene function prediction problem. It aims to predict the
association probability between each gene and function by taking
advantage of the GCN spectral decomposition, the information available
of associations between genes and functions, and the ancestral
relations between the functions (i.e., the GO hierarchy).

A case study on \textit{Zea mayz} (maize) is presented. Using the
structural information of the gene co-expression network (extracted by
a spectral clustering algorithm) and considering the hierarchical
structure of the biological processes (using HMC) seems to be the key
for the improved performance of the proposed approach. More precisely,
the global HMC method, which considers all features available for a
sub-hierarchy to build a single classifier, outperforms the other
methods in relation to the three metrics that were used (namely,
AU($\overline{\textnormal{PRC}}$) , $\overline{\textnormal{AUPRC}}$ ,
and $\overline{\textnormal{AUPRC}_w}$).

The results presented in~\cite{romero-clust-2022} show that the
features extracted from the GCN using spectral clustering lead to
better prediction performance in the gene function prediction task
(addressed as an independent binary classification problem per
function). In this work, it has been shown that considering the
ancestral relations between functions to produce an outcome that
satisfies its hierarchical structure (i.e., complies the true-path
rule or hierarchical constraint), based on the features extracted from
the GCN, improves the performance in the gene function prediction task
(addressed as a hierarchical multi-label classification problem).

Two main lines of work can be considered for future work.  First,
applying the proposed approach to identify genes associated to
specific stresses (e.g., low temperature, salinity) can help to reduce
the set of candidate genes that respond to treatments for in vivo
validation.  Second, exploring transfer learning techniques
(especially, domain adaptation) to enrich the building of the
classifiers using information from other organisms (datasets), not
only can lead to higher prediction performance, but also can enable
the proposed approach on organisms without a wealth of significant
functional information.

\section*{Abbreviations}
\begin{description}
	\item[AUPRC:] Area Under Precision-Recall Curve
	\item[DAG:] Directed Acyclic Graph
	\item[GO:] Gene Ontology 
	\item[GCN:] Gene Co-expression Network
	\item[HMC:] Hierarchical Multi-label Classification
	\item[\textit{lcl}:] Local Classifier per Level
	\item[\textit{lcn}:] Local Classifier per Node
	\item[\textit{lcpn}:] Local Classifier per Parent Node
\end{description}

\section*{Availability of data and materials}
The datasets analyzed for the current study are publicly available from
different sources. They can be found in the following locations:

\begin{itemize}	
	\item Gene co-expression data of \textit{Oryza sativa Japonica} is
	available on ATTED-II~\cite{obayashi-atted2018-2018}.
	\item Functional data of rice genes is available on the DAVID Bioinformatics Resources~\cite{huang-david-2009}.
	\item Hierarchical data of Gene Ontology terms are available on the GOATOOLS Python library~\cite{klopfenstein-goatools-2018}.
\end{itemize}

The data collected, cleaned, and processed from the above sources as
used in the case study can be requested to the authors.

A workflow implementation is publicly available:
\begin{itemize}
	\item Project name: clustering\_hmc
	\item Project home page: \url{https://github.com/migueleci/clustering_hmc}
	\item Operating system(s): platform independent.
	\item Programming language: Python 3.
	\item Other requirements: None.
	\item License: GNU GPL v3.
\end{itemize}

\section*{Funding}
This work was partially funded by the OMICAS program: Optimización
Multiescala In-silico de Cultivos Agrícolas Sostenibles
(Infraestructura y Validación en Arroz y Caña de Azúcar), anchored
at the Pontificia Universidad Javeriana in Cali and funded within
the Colombian Scientific Ecosystem by The World Bank, the Colombian
Ministry of Science, Technology and Innovation, the Colombian
Ministry of Education and the Colombian Ministry of Industry and
Turism, and ICETEX, under GRANT ID: FP44842-217-2018. The second
author was partially supported by Fundación CeiBA.

\bibliographystyle{abbrv}
\bibliography{main}

\end{document}